\documentclass[10pt,twocolumn,letterpaper]{article}

\usepackage[pagenumbers]{wacv}              % To produce the CAMERA-READY version
\usepackage{graphicx}
\usepackage{amsmath}
\usepackage{amssymb}
\usepackage{booktabs}
\usepackage{multirow}
\usepackage{caption}
\usepackage{hyperref}

\usepackage[capitalize]{cleveref}
\crefname{section}{Sec.}{Secs.}
\Crefname{section}{Section}{Sections}
\Crefname{table}{Table}{Tables}
\crefname{table}{Tab.}{Tabs.}

\def\wacvPaperID{37} % *** Enter the WACV Paper ID here
\def\confName{WACV}
\def\confYear{2025}
\begin{document}

\title{HipyrNet: Hypernet-Guided Feature Pyramid network for mixed-exposure correction}
\author{
Shaurya Rathore*, Aravind Shenoy*, Krish Didwania, Aditya Kasliwal, Ujjwal Verma\\
Manipal Institute of Technology, MAHE\\
Manipal, India\\
{\tt\small shauryarathore121@gmail.com, aravindshenoywork@gmail.com}
}
\maketitle

\maketitle
\begin{abstract}
Recent advancements in image translation for enhancing mixed-exposure images have demonstrated the transformative potential of deep learning algorithms. However, addressing extreme exposure variations in images remains a significant challenge due to the inherent complexity and contrast inconsistencies across regions. Current methods often struggle to adapt effectively to these variations, resulting in suboptimal performance. In this work, we propose HipyrNet, a novel approach that integrates a HyperNetwork within a Laplacian Pyramid-based framework to tackle the challenges of mixed-exposure image enhancement. The inclusion of a HyperNetwork allows the model to adapt to these exposure variations. HyperNetworks dynamically generates weights for another network, allowing dynamic changes during deployment. In our model, the HyperNetwork employed is used to predict optimal kernels for Feature Pyramid decomposition, which enables a tailored and adaptive decomposition process for each input image. Our enhanced translational network incorporates multiscale decomposition and reconstruction, leveraging dynamic kernel prediction to capture and manipulate features across varying scales. Extensive experiments demonstrate that HipyrNet outperforms existing methods, particularly in scenarios with extreme exposure variations, achieving superior results in both qualitative and quantitative evaluations. Our approach sets a new benchmark for mixed-exposure image enhancement, paving the way for future research in adaptive image translation.
\end{abstract}
\def\thefootnote{*}\footnotetext{Both authors contributed equally to this work}\def\thefootnote{\arabic{footnote}}
\def\thefootnote{}\footnotetext{Accepted at WACV 2025 Workshop - 4th Workshop on Image/Video/Audio Quality in Computer Vision and Generative AI}\def\thefootnote{\arabic{footnote}}

\section{Introduction}

Images captured under non-ideal illumination conditions, such as underexposure and overexposure, often suffer from significant color distortion, loss of detail, and compromised aesthetic quality. These issues not only degrade the image's visual appeal but also complicate tasks such as object recognition and localization. To address these problems, various exposure correction techniques have been developed to improve and restore details in both bright and dark areas of an image. These techniques aim to recover the original content and enhance the overall aesthetics of the image.

One of the primary challenges in exposure correction is that underexposure or overexposure requires distinct approaches. Underexposed images typically need techniques that involve brightening the image and reducing noise. On the other hand, overexposed images necessitate retrieving lost highlight information without introducing artifacts or unnatural colors. Given these contrasting needs, a single algorithm that can effectively correct underexposure and overexposure is challenging to develop and implement ~\cite{Guo2020ZeroReferenceDC}~\cite{Huang_2023_CVPR}.

Traditionally, exposure correction methods have focused on low-light image enhancement (LLIE) ~\cite{7782813}~\cite{Lore2015LLNetAD}~\cite{Lv2018MBLLENLI}, primarily enhancing images captured under low-light or dark conditions. These methodologies train on low-light pictures and aim to enhance the brightness and contrast of these images. This approach has been successful in improving the visibility of details in dark regions, where fine-grained textures are often difficult to distinguish due to the lack of light. However, LLIE methods generally struggle to deal with images that feature regions with uneven lighting ~\cite{Guo2020ZeroReferenceDC}~\cite{BenchmarkLLIE}, where portions of the image are underexposed while others are overexposed.

Mixed exposure, as the name suggests, refers to images where both underexposed and overexposed regions coexist within the same frame. These images present a unique challenge because achieving a balanced exposure requires simultaneously addressing the dark details in underexposed areas and the bright highlights in overexposed regions. Unlike traditional Low-Light Image Enhancement (LLIE), which primarily focuses on brightening dark areas while preserving details, mixed exposure correction involves an intricate process of adjusting exposure levels across the entire image. This added complexity arises from the need to ensure that both extremes of exposure are harmonized effectively without introducing artifacts, such as halo effects or unnatural tonal transitions. Consequently, mixed exposure correction is a significantly more challenging task, requiring more advanced and adaptive algorithms to manage such highly dynamic variations.

In recent years, there has been increasing interest in using image-to-image translation (I2IT) methods to address the exposure correction task, particularly through encoding-decoding architectures ~\cite{Lore2015LLNetAD}~\cite{Isola_2017_CVPR}~\cite{Wang_2018_CVPR}. The use of I2IT models is particularly appealing because of their ability to learn complex mappings between images. By training the models to recognize and correct the variations in exposure, I2IT methods can effectively restore image details and improve visual quality. However, these methods are often limited to low-resolution tasks or rely on computationally intensive 
models, making them impractical for real-world applications. Additionally, while these models show promise for correcting individual exposure problems, they are often not suitable for handling mixed exposure, where both overexposed and underexposed regions coexist.

Many algorithms attempt to address the mixed-exposure task by using algorithms designed for single-exposure correction. While most of these algorithms effectively handle single-exposure images, they often rely on conventional methods that utilize manually designed strategies ~\cite{4266947}~\cite{4895264}~\cite{Fu_2016_CVPR} and deep-learning-driven approaches that leverage complex neural networks ~\cite{wei2018deepretinexdecompositionlowlight}~\cite{zhang2019kindlingdarknesspracticallowlight}~\cite{Guo2020ZeroReferenceDC} that tend to struggle when dealing with mixed-exposure images. The problem lies in the fact that current algorithms generally excel at correcting either one of these exposure types but not both simultaneously. A method that can handle both types of exposure correction without compromising image quality across the entire frame is crucial for mixed exposure tasks.

In this paper, we propose a method to improve multiple-exposure correction by introducing a dynamic adaptation mechanism for the network. Our method leverages a Hypernet module designed to tailor the network's response to each input image, enabling real-time adaptivity during inference. We train our model on a comprehensive dataset with both underexposed and overexposed images and evaluate it on a challenging mixed-exposure dataset, demonstrating the generalization capability of our model.

Particularly, the main contributions of this work are summarized as:
\begin{itemize}
    \item We propose a novel Hypernet-driven approach that dynamically predicts weights for the decomposition kernel of Laplacian Pyramids based on input image characteristics- allowing the network to adjust its parameters in real-time for optimal exposure correction.
    %\item We incorporate the Laplacian Pyramid Translation Network proposed by Liang et al. ~\cite{liang2021highresolutionphotorealisticimagetranslation} and implement a customized residual block structure, enabling it to perform effective exposure correction.
    \item We combine classical computer vision techniques, such as Laplacian Pyramids, and modern Deep Learning methods, such as feature extraction, to derive a novel architecture inspired by 
    % the Laplacian Pyramid Translation Network proposed by Liang et al. 
    ~\cite{liang2021highresolutionphotorealisticimagetranslation}.
    \item  Our method is extensively evaluated on challenging datasets (SICE Grad and SICE Mix ~\cite{zheng2024lowlightimagevideoenhancement}) and demonstrates state-of-the-art results across PSNR and SSIM, exemplifying our model's robustness and real-world use.
\end{itemize}

\section{Related Works}
\subsection{Image contrast Enhancement}

Traditional low-light image enhancement (LLIE) techniques often rely on fundamental approaches such as histogram equalization ~\cite{4895264} and Retinex theory. Retinex-based approaches aim to decompose an image into its illumination and reflectance components, where the reflectance captures the intrinsic scene properties, and the illumination represents the lighting conditions ~\cite{wei2018deepretinexdecompositionlowlight} method. By focusing on enhancing the reflectance component, these methods improve image contrast while preserving a natural appearance. This makes them particularly effective for addressing challenges posed by uneven lighting and shadowed regions. However, many existing methods struggle to adequately suppress noise artifacts and are computationally intensive to train, limiting their practical efficiency.

Deep learning architectures, including Convolutional Neural Networks (CNNs) ~\cite{wang2019progressiveretinexmutuallyreinforced}~\cite{8305143} and other advanced models, have demonstrated the ability to learn complex mappings between low-quality and high-quality images directly from data. Numerous studies have explored deep learning-based approaches to tackle the challenges associated with exposure correction, particularly the problems of image color shifts ~\cite{yan2024needcolorspaceefficient} and noise stabilization. Works in this field have experimented with feature pyramid networks for extracting the latent representation of improperly exposed images ~\cite{10274690}. These methods have been effective in achieving state-of-the-art results on benchmark datasets and have shown notable success when addressing underexposure and overexposure as separate tasks. In this paper, we address the challenge of enhancing complex datasets comprising images with both underexposed and overexposed regions in the same image, such as the SICE Grad and SICE Mix datasets ~\cite{zheng2024lowlightimagevideoenhancement}. We aim to develop a generalized model capable of performing robustly across a diverse range of exposure conditions, as exemplified by our work on the SICE dataset ~\cite{8259342}.
% Furthermore, recent advancements in this domain have extended to more challenging scenarios involving mixed exposure correction.

Existing methodologies predominantly rely on the use of illumination maps ~\cite{adhikarla2024unifiedegformerexposureguidedlightweight} and modifications to the color space ~\cite{yan2024needcolorspaceefficient} representation of images. While these techniques have achieved significant progress in improving the overall visual quality of images, they are not without limitations. Specifically, these approaches are often prone to producing over-smoothed outputs, which can result in a loss of fine-grained image details. Additionally, the inherent transformations involved in these methods frequently introduce visual artifacts, further compromising the natural appearance of the enhanced images. Consequently, there remains a need for more robust and perceptually consistent solutions to address these limitations effectively.

\subsection{Feature Pyramid Networks}
\begin{figure*}[ht!]
    \centering
    \includegraphics[width=0.95\textwidth]{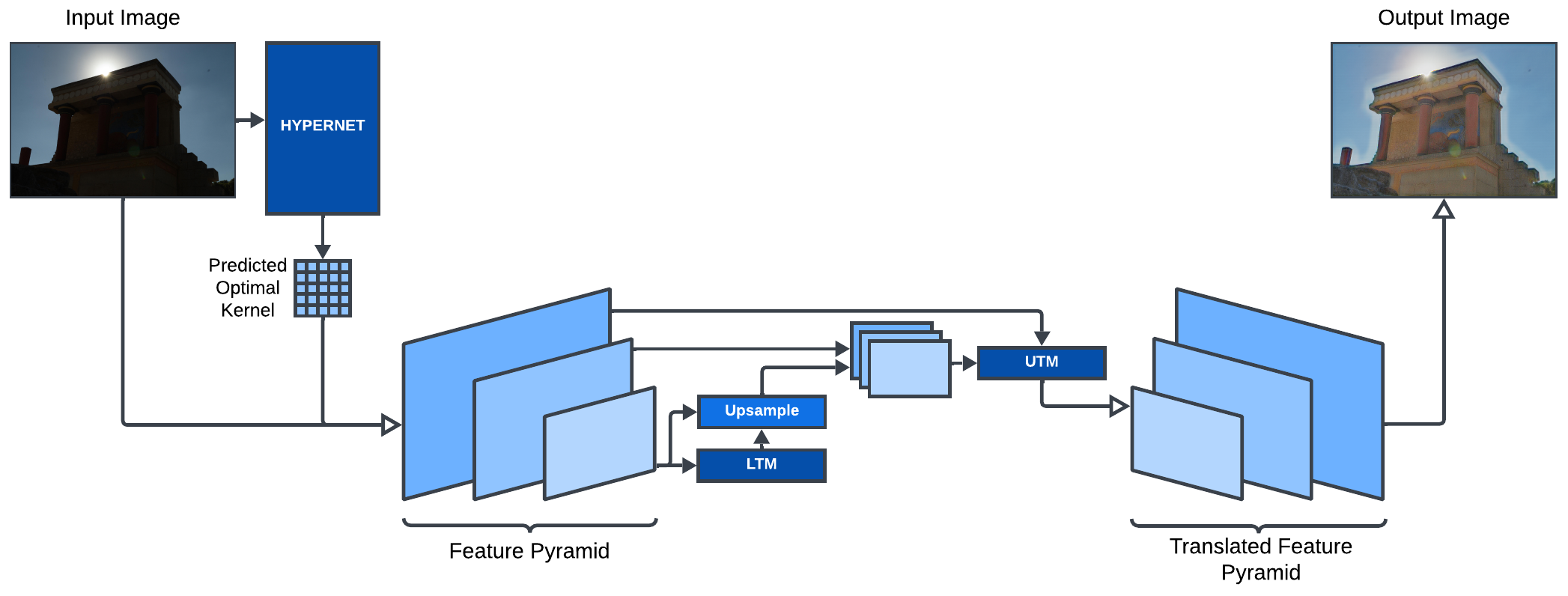}
    % \caption{Model architecture of the proposed HipyrNet exemplified with input from the SICE Dataset}
    \caption{A schematic of the proposed model. The figure represents the Hypernet, and the following model uses the predicted kernel to translate the underexposed image. UTM (Upper Translational Module) and LTM (Lower Translational Module) translate the feature pyramid. The exact architecture of the Hypernet is denoted in Fig. \ref{fig:hypernet}. The image is taken from the SICE Dataset ~\cite{8259342}.}
    \label{fig:model}
\end{figure*}

Feature Pyramid Networks (FPNs) have become a cornerstone in modern computer vision tasks, particularly for object detection, semantic segmentation, and image restoration. Introduced by Lin et al. in 2017 ~\cite{lin2017featurepyramidnetworksobject}, FPNs utilize a hierarchical top-down architecture to construct a feature pyramid. By leveraging the hierarchical feature maps generated by convolutional networks, FPNs enable robust detection of objects at various scales, making them highly effective for tasks requiring multi-level feature integration.

A key strength of FPNs lies in their ability to handle multiscale features. This approach has been successfully adopted in numerous frameworks, including Mask R-CNN ~\cite{he2018maskrcnn} for instance, segmentation and RetinaNet ~\cite{lin2018focallossdenseobject} for anchor-based object detection. FPNs have demonstrated significantly improved accuracy and efficiency compared to earlier methods, which often struggled to balance feature resolution and computational cost.

\subsection{Laplacian Pyramid Translation Network}
The Laplacian Pyramid Translation Network (LPTN) ~\cite{liang2021highresolutionphotorealisticimagetranslation} is an efficient and scalable deep learning framework for image enhancement tasks, including super-resolution, low-light image enhancement, and other image-to-image translation problems. It leverages the Laplacian Pyramid, a hierarchical structure that decomposes an image into frequency bands, enabling simultaneous enhancement of global structures (low-frequency components) and fine details (high-frequency components).

LPTN’s architecture follows a coarse-to-fine approach. The input image is decomposed into a low-resolution base, representing global features like brightness and contrast and high-frequency residuals, capturing textures and sharpness. These components are processed separately by lightweight neural networks specialized for each frequency band.

The high-level framework focuses on global adjustments such as brightness, contrast, and color consistency, while the networks handling the high-frequency residuals refine intricate textures and fine details. After processing, the image is reconstructed by iteratively adding the enhanced high-frequency details to the upsampled base image, ensuring that both global and local features are preserved.

This multiscale processing strategy enables LPTN to achieve high-quality results with significantly reduced memory due to the non-parametric pyramid decomposition and lesser computational costs due to the focus of intensive operations on smaller, downsampled images. By addressing both global and local features effectively, LPTN overcomes the limitations of traditional CNNs, which often struggle to handle high-resolution inputs or balance global corrections with local detail preservation.

Additionally, LPTN’s modular and flexible design makes it adaptable to a variety of image enhancement tasks. Its ability to process high-resolution images with limited computational resources makes it particularly suited for real-world applications where efficiency and scalability are critical. 

While LPTN is efficient and effective for image restoration, it struggles with complex degradations like mixed exposure images. Its local focus and static feature handling limit its ability to adapt to varying exposures and capture global dependencies, leading to suboptimal performance on challenging datasets and real-world scenarios.

Our proposed model builds upon LPTN’s lightweight architecture but introduces hypernetworks to dynamically adjust the network parameters based on the input image characteristics. This enables the model to adapt flexibly to varying exposure conditions, making it well-suited for images containing both underexposed and overexposed areas.

\subsection{HyperNetworks}
 HyperNetworks (HNs) ~\cite{ha2016hypernetworks} are a class of deep neural networks designed to generate the weights of another target network responsible for performing a specific learning task.
 
 Unlike traditional neural networks with fixed weights, HNs dynamically predict the weights based on the input they receive, enabling them to adapt to diverse conditions or tasks. 

HNs have been successfully applied in various domains of machine learning, including language modeling, computer vision, continual learning, hyperparameter optimization, multi-objective optimization, and decoding block codes. One of their key strengths lies in generating personalized models, as they can create target networks conditioned on input data, making them particularly effective for applications requiring adaptability.

\section{Architecture}
Fig. \ref{fig:model} provides a summarized view of the architecture, which begins by processing an input image with a HyperNetwork, which returns a personalized kernel tailored to the image's specific characteristics. This dynamically generated kernel is then used for the pyramid decomposition. The components of the resulting feature pyramid are then processed through translational modules. The lower pyramid level is refined and translated through the Lower Translational Module (LTM). The output of the LTM, along with the upper levels of the pyramid, is passed through the Upper Translational Module (UTM), which returns the Translated Feature Pyramid. The translated pyramid is then reconstructed back into the output image by upsampling and adding its components.

\subsection{Feature Pyramid}

The Laplacian Pyramid is a classical technique in image processing that linearly decomposes an image into a series of high- and low-frequency components, enabling the exact reconstruction of the original image in a non-parametric manner. This principle motivates us to integrate a personalized approach to sample-specific processing by predicting the weights of the kernel used in the decomposition of LP through the Hypernet. Mathematically, the modified Laplacian Pyramid process can be described as follows:

Given an input image \(I_0 \in \mathbb{R}^{H \times W}\), the first step is to calculate a low-pass prediction \(I_1 \in \mathbb{R}^{\frac{H}{2} \times \frac{W}{2}}\) by applying a convolution operation with the dynamically generated kernel \(K_{\text{image}}\). Each pixel in \(I_1\) is computed as a weighted sum of neighboring pixels in \(I_0\):

 \begin{equation}
     I_1(x, y) = \sum_{(u, v) \in \mathcal{N}(x, y)} K_{\text{image}}(u, v) \cdot I_0(u, v)
 \end{equation}

where \(\mathcal{N}(x, y)\) denotes the neighborhood around pixel \((x, y)\) in \(I_0\) and \(K_{\text{image}}(u, v)\) is the kernel weight. In our work, we utilize a dynamically generated kernel, as detailed previously.

To ensure reversible reconstruction, the high-frequency residual \(h_0\) is computed as:

\begin{equation}
    h_0 = I_0 - \text{Upsample}(I_1)
\end{equation}

where the \(\text{Upsample}(\cdot)\) function increases the resolution of \(I_1\) using bicubic interpolation. The process is iteratively applied to \(I_1\) to decompose the image into subsequent levels further, producing a sequence of low- and high-frequency components.

The hierarchical decomposition results in multiple pyramid levels. In our architecture, we employ three levels, corresponding to resolutions \(H \times W\), \(\frac{H}{2} \times \frac{W}{2}\), and \(\frac{H}{4} \times \frac{W}{4}\). Mathematically, the decomposition can be represented as:

\begin{equation}
\begin{aligned}
I_1 & = \text{Downsample}(I_0), \quad h_0 = I_0 - \text{Upsample}(I_1) \\
I_2 & = \text{Downsample}(I_1), \quad h_1 = I_1 - \text{Upsample}(I_2) \\
I_3 & = \text{Downsample}(I_2),  \quad h_2 = I_3\\
\end{aligned}
\end{equation}

The final pyramid consists of \(\{h_0\}\),\(\{h_1\}\) and \(\{h_2\}\).

\subsection{Loss Function}
The proposed method incorporates a multicomponent loss function to train the generator effectively. The total generator loss (\(L_G\)) is composed of three key components: pixel reconstruction loss, adversarial loss, and kernel loss (if a hypernetwork is used). These components are designed to jointly optimize the model for accurate image restoration, adversarial robustness, and kernel prediction.

1. \textbf{Pixel Reconstruction Loss (\(L_{\text{pix}}\))}: This loss is calculated between the generated output (\(\hat{y}\)) and the high-light image(HLI) using Mean Squared Error (MSE):
\begin{equation}
   L_{\text{pix}} = \text{MSE}(\hat{y}, \text{HLI})
\end{equation}

2. \textbf{Adversarial Loss (\(L_{\text{GAN}}\))}: This loss is calculated using the output of the discriminator (\(D(\hat{y})\)) and the generated image basis the gan type used.

3. \textbf{Kernel Loss (\(L_{\text{ker}}\))}: For models utilizing a hypernetwork, this loss is calculated between the predicted kernel (\(\hat{K}\)) and the target kernel (\(K_{\text{opt}}\)) using Mean Squared Error (MSE):

\begin{equation}     
L_{\text{ker}} = \text{MSE}(\hat{K}, K_{\text{opt}})
\end{equation}

The target kernel here refers to the Gaussian Kernel used in the LPTN framework~\cite{liang2021highresolutionphotorealisticimagetranslation}. The loss is used to update the model, allowing the Hypernet-predicted kernel to be close to the Gaussian kernel but also be dynamic to the input.

The final generator loss is defined as:
\begin{equation}
L_G = L_{\text{GAN}} + \eta_{\text{pix}} L_{\text{pix}} +  \lambda_{\text{ker}} L_{\text{ker}}
\end{equation}
where \(\lambda_{\text{ker}}\) controls the contribution of the kernel loss and \(\eta_{\text{pix}}\) controls the contribution of the pixel-wise loss to the total loss. The loss is then used to optimize the generator and hypernetwork (if applicable), while the discriminator weights are frozen during generator updates.

This composite loss function balances reconstruction accuracy, perceptual quality, and adaptability to varying degradations, enabling robust performance across diverse exposure conditions.

\subsection{HyperNet Architecture}
In our work, we propose a HyperNetwork that is employed to predict a personalized kernel for each image, tailored for its decomposition in the Feature Pyramid. Given the inherent variations in images, a single fixed kernel might not be optimal for all cases. Instead, the HyperNetwork generates a kernel that accounts for the specific characteristics of each input image.

\begin{figure}[h!]
    \centering
2    \includegraphics[width=0.45\textwidth]{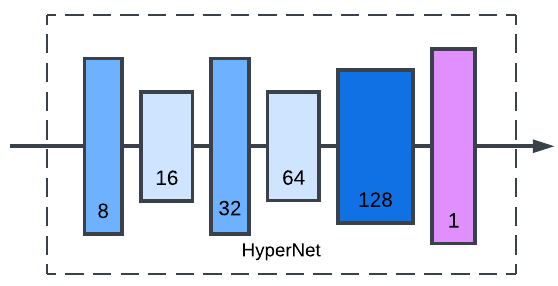}
    \caption{Expanded Schematic of the Hypernet. The figure represents the Hypernet architecture in detail and shows each of the convolution blocks and their corresponding output channels}
    \label{fig:hypernet}
\end{figure}

The exact \textit{HyperNet} architecture as shown in \ref{fig:hypernet} is designed to dynamically produce a kernel of dimensions \(5 \times 5\). The network consists of six convolutional layers with progressively increasing channels and employs irregular convolutional kernels and pooling operations to achieve significant downsampling. The architectural details are as follows:

\begin{enumerate}
    \item The initial convolutional layer uses an irregular kernel of size \(3 \times 1\), followed by a ReLU activation and a max-pool operation.
    \item The second convolutional layer utilizes a \(1 \times 3\) kernel, followed by a ReLU activation and max pooling, continuing the downsampling process.
    \item The third and fourth convolutional layers alternate between \(3 \times 1\) and \(1 \times 3\) kernels, respectively, each followed by ReLU activations and pooling layers.
    \item The fifth convolutional layer uses a \(3 \times 3\) followed by max pooling and a final \(1 \times 1\) convolution to precisely reduce the dimensions to \(5 \times 5\).
\end{enumerate}

The use of irregular kernels (\(3 \times 1\) and \(1 \times 3\)) is motivated by the need to balance computational efficiency and spatial feature extraction. These kernels enable the model to focus independently on horizontal and vertical gradients, capturing fine-grained details along each axis without the overhead of larger square kernels. Furthermore, irregular kernels introduce an asymmetric receptive field, which is particularly advantageous for processing images with anisotropic features or dominant patterns in one direction.

By alternating irregular and square kernels, the architecture ensures the ability to capture both local and global spatial dependencies. 
% This lightweight yet effective design achieves computational efficiency while adapting to diverse inputs, making it well-suited for dynamic ; thel generation.
\begin{figure*}[t]
    \centering
    \includegraphics[width=0.9\textwidth]{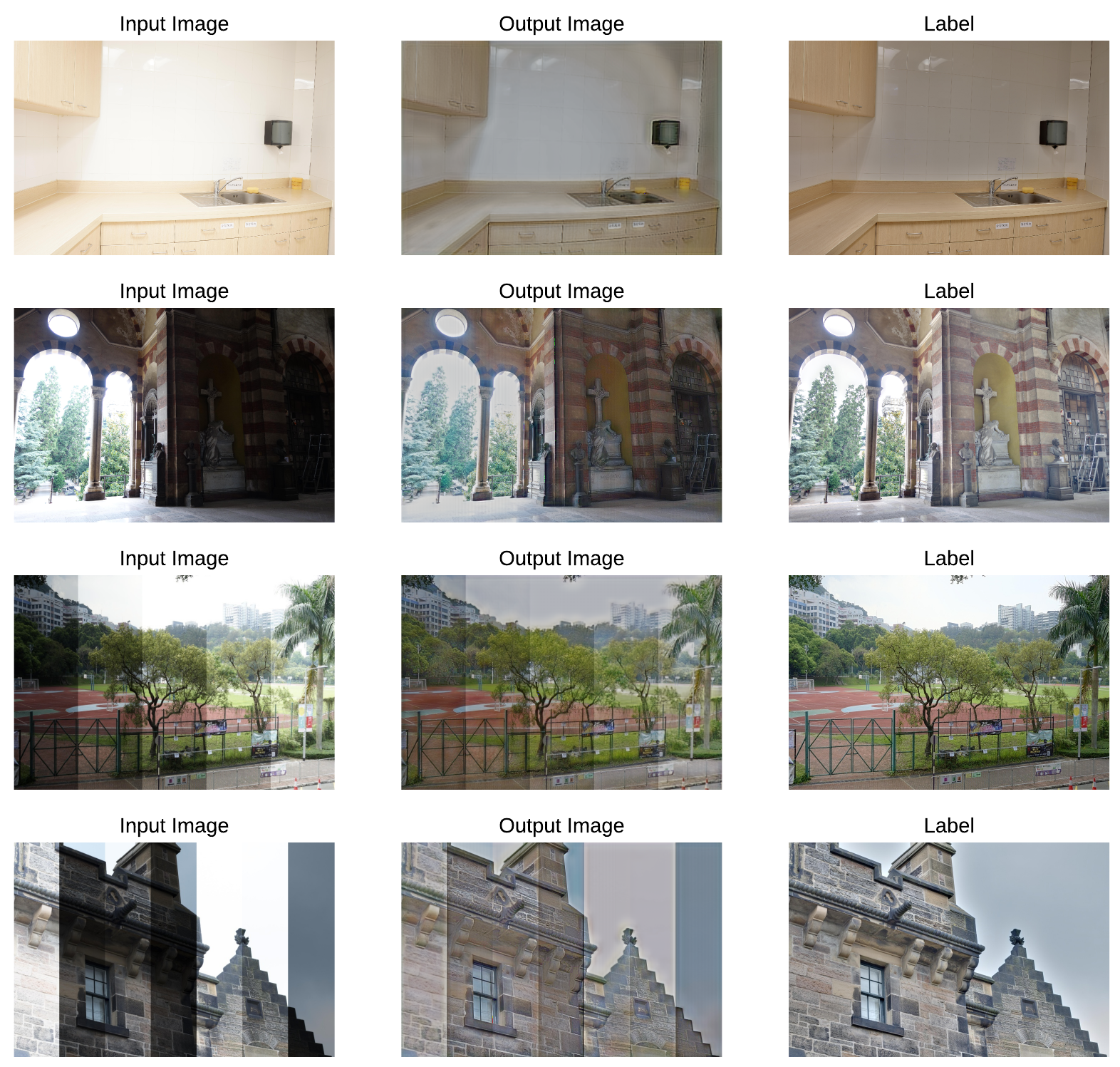} 
    \caption{Visualizations of images from the SICE ~\cite{8259342} and the SICE Grad and SICE Mix ~\cite{zheng2024lowlightimagevideoenhancement} Dataset. The first row refers to an overexposed image from SICE, the second row is of an underexposed image from SICE, and the last two rows refer to SICE Grad and SICE Mi,x respectively}
    \label{fig:sice_visual}
\end{figure*}
\section{Experiments}

\subsection{Datasets}
In our study, we employ 4 datasets to rigorously train and evaluate our proposed model: SICE-v1 and SICE-v2 for foundational training, including the SICE-Grad and SICE-Mix subsets, which address gradient challenges and mixed-exposure challenges, respectively, for testing purposes.
The SICE dataset~\cite{8259342} contains a total of 589 sets of images, with each set having 7 or 9 images of varying contrast from overexposed to underexposed. The training set, validation set, and test set are divided into three groups in a split of 8:1:1. The test set was defined by the testing indexes provided in the dataset.

The SICE Grad and SICE Mix ~\cite{zheng2024lowlightimagevideoenhancement} datasets are derived from the original SICE dataset.
For SICE Grad, Panels from images of SICE with different exposures were selected and arranged such that the exposure increases from left to right. On the other hand, the SICE Mix dataset was constructed by permuting these panels. These derivations provide benchmarks tailored for testing model performance on uneven and mixed exposure challenges.

We train on the SICE training set and test on SICE, SICE-Mix and SICE-Grad. We resize the original SICE image to 608x896 which is the same for SICE-Grad or SICE-Mix images.

\subsection{Hyperparameter Tuning}
During training, we use the Adam optimizer ~\cite{adam} with a learning rate of \( 10^{-4} \). Stochastic Gradient Descent (SGD) is employed for the training process. All methods are evaluated in terms of PSNR and SSIM.
Experiments conducted on the number of residual block yield as shown in Table \ref{tab:nrb}, the optimal configuration which comprises of two residual blocks in the Lower Translational Module(LTM) and four blocks in the Upper Translational Module(UTM).

\begin{table}[ht]
\centering
\renewcommand{\arraystretch}{1.2}
\begin{tabular}{c c c c}
\hline
\textbf{UTM} & \textbf{LTM} & \textbf{PSNR↑} & \textbf{SSIM↑} \\ \hline
2 & 3       & 20.592 & 0.73956 \\
2 & 4              & \textbf{20.845} & \textbf{0.74675} \\
3 & 2              & 20.307 & 0.74669 \\
4 & 2              & 20.434 & 0.75191 \\ \hline
\end{tabular}
\caption{Effect of number of residual blocks on PSNR and SSIM in the Upper Translational Module and the Lower Translational Module. Higher scores are better}
\label{tab:nrb}
\end{table}

It can be seen that increasing the number of residual blocks in the LTM branch improved PSNR, while a similar increase in the UTM branch enhanced SSIM. The UTM branch focuses on higher-frequency features, such as edges, while the LTM branch captures lower-frequency features. This indicates that higher-frequency features, which are more closely associated with structural information, contribute to improvements in SSIM. In contrast, lower-frequency features primarily enhance visual quality, resulting in higher PSNR.

\subsection{Gan Loss Functions}
Various experiments were conducted to evaluate the performance of the loss function, focusing on different loss weights and GAN types.
The optimal loss weight for the MSE was found to be 1000 and 0.01 for the Kernel loss Weight.
The different gan types used were Vanilla GAN loss being the original GAN (BCE)  ~\cite{GAN}, LS-GAN loss(Least Squares GAN)  ~\cite{LSGAN}, which minimizes the difference between the discriminator's output and a target value, WGAN ~\cite{WGAN} and WGAN-Softplus ~\cite{WGANSP} (Wasserstein GAN) loss which uses the Wasserstein distance between the real and generated images and Hinge loss ~\cite{lim2017geometricgan}   which aims to generate image that push the discriminators score above the defined Margin.

After a comprehensive evaluation, the best results are achieved using the GAN type WGAN Softplus.  

\begin{table}[ht]
\centering
\begin{tabular}{l c c}
\toprule
\textbf{GAN Loss Function} & \textbf{PSNR↑} & \textbf{SSIM↑} \\
\midrule
Vanilla ~\cite{GAN}           & 21.020 & 0.76841 \\
LS-GAN ~\cite{LSGAN}            & 21.059 & 0.74214 \\
W-GAN ~\cite{WGAN}             & 20.985 & 0.74668 \\
W-GAN Softplus ~\cite{WGANSP}   & \textbf{21.081} & \textbf{0.76801} \\
Hinge ~\cite{lim2017geometricgan}             & 20.845 & 0.74675 \\
\bottomrule
\end{tabular}
\caption{Effect of different types of GANs used in the GAN loss function on PSNR and SSIM. }
\label{tab:gan_type}
\end{table}

\section{Results}

\begin{table}[ht]
\centering
\setlength{\tabcolsep}{6pt} % Adjust column spacing
\renewcommand{\arraystretch}{1.1} % Adjust row spacing
\begin{tabular}{llcc}
\toprule
\textbf{Dataset}           & \textbf{Method} & \textbf{PSNR↑} & \textbf{SSIM↑} \\
\midrule
\multirow{2}{*}{\textbf{SICE}} 
                           & LPTN            & 18.316         & 0.697          \\
                           & HipyrNet      & \textbf{21.081}         & \textbf{0.768}          \\
\hline
\multirow{2}{*}{\textbf{SICE Grad}} 
                           & LPTN            & 16.201         & 0.633          \\
                           & HipyrNet        & \textbf{16.324}         & \textbf{0.710}          \\
\hline
\multirow{2}{*}{\textbf{SICE Mix}}  
                           & LPTN            & 15.515         & 0.600          \\
                           & HipyrNet        & \textbf{15.857}         &\textbf{ 0.668}          \\
\bottomrule
\end{tabular}
\caption{Performance metrics for LPTN ~\cite{liang2021highresolutionphotorealisticimagetranslation} and HipyrNet on SICE ~\cite{8259342}, SICE Grad, and SICE Mix ~\cite{zheng2024lowlightimagevideoenhancement}.}
\label{tab:sice_results}
\end{table}

As shown in Table \ref{tab:sice_results}, there is an increase in both PSNR and SSIM as compared to LPTN. The results indicate that the inclusion of the HyperNetwork makes the model adapt to the inputs, allowing for dynamic pyramid decomposition and improving the overall visual quality of the output.

\begin{table}[ht]
\centering
\setlength{\tabcolsep}{4pt} % Reduce column spacing
\renewcommand{\arraystretch}{1.1} % Adjust row height
\resizebox{\linewidth}{!}{ % Automatically adjust to page width
\begin{tabular}{l c c c c}
\hline
\multirow{2}{*}{\textbf{Methods}} & \multicolumn{2}{c}{\textbf{SICE Grad}} & \multicolumn{2}{c}{\textbf{SICE Mix}} \\
                                   & \textbf{PSNR ↑} & \textbf{SSIM ↑}      & \textbf{PSNR ↑} & \textbf{SSIM ↑}      \\ \hline
\multirow{1}{*}{RetinexNet ~\cite{wei2018deepretinexdecompositionlowlight}}        & 12.397          & 0.606               & 12.450          & 0.619               \\
ZeroDCE ~\cite{Guo2020ZeroReferenceDC}                           & 12.428          & 0.633               & 12.475          & 0.644               \\
RAUS ~\cite{BenchmarkLLIE}                               & 0.864           & 0.493               & 0.868           & 0.494               \\
SGZ ~\cite{sgz}                                & 10.866          & 0.607               & 10.870          & 0.607               \\
LLFlow ~\cite{llienf}                            & 12.737          & 0.617               & 12.737          & 0.617               \\
URetinexNet ~\cite{uretinex}                       & 10.903          & 0.600               & 10.894          & 0.602               \\
SCI ~\cite{ma2022fastflexiblerobustlowlight}                                & 8.644           & 0.529               & 8.625           & 0.531               \\
KinD ~\cite{BeyondBL}                              & 12.986          & 0.656               & 13.144          & 0.668               \\
KinD++ ~\cite{BeyondBL}                             & 13.196          & 0.663               & 13.346          & 0.680               \\
U-EGformer ~\cite{adhikarla2024unifiedegformerexposureguidedlightweight}                         & 13.272          & 0.643               & 14.235          & 0.652               \\
\textbf{HyperLPTN}                 & \textbf{16.602} & \textbf{0.709}       & \textbf{16.158} & \textbf{0.688}       \\ \hline
\end{tabular}
} % End resizebox
\caption{Comparison of existing methods against ours across PSNR and SSIM on the SICE Mix and SICE Grad ~\cite{zheng2024lowlightimagevideoenhancement} datasets. Our model was trained on SICE and evaluated across SICE Grad and SICE Mix.}
\label{tab:sicegm_results}
\end{table}

In Table~\ref{tab:sicegm_results}, we compare state of the art models with our model where we achieve state-of-the-art results on the SICE Grad and SICE Mix datasets. HipyrNet shows a 25\% improvement and a 13.5\% increase in PSNR on the SICE Grad and the SICE Mix dataset, respectively.
For visualizations, one overexposure and one underexposure image from the SICE dataset have been chosen along with visual results for the SICE Grad and SICE Mix datasets in Fig.~\ref{fig:sice_visual}.

\section*{Conclusion and Future Work}

In summary, we present \textbf{HipyrNet}, an approach that integrates HyperNetworks into Laplacian Pyramid-based Translation Networks for improved kernel prediction. By dynamically generating personalized kernels for each input image, HipyrNet significantly enhances contrast and detail in mixed-exposure images, leading to notable improvements in network performance and image quality across diverse datasets. This approach effectively leverages the adaptability of HyperNetworks to optimize the decomposition process, ensuring robust performance across varying conditions. Future work includes extending this methodology to other applications such as super-resolution, image generation, and semantic segmentation. Additionally, exploring its integration into real-time video processing systems could open new possibilities for enhancing visual fidelity in challenging scenarios. HipyrNet provides a strong foundation for further research in adaptive and personalized image processing techniques.

\section*{Acknowledgements}
We would like to thank Mars Rover Manipal, an interdisciplinary student project of MAHE, for providing the essential resources and infrastructure that supported our research. We also extend our gratitude to Ishaan Ghakar for his valuable guidance in reviewing and refining the manuscript, and to Mohammed Sulaiman for his contributions in facilitating access to additional resources crucial to this work.  

\bibliographystyle{ieee_fullname}
\bibliography{hyperlptn1}
\end{document}